\documentclass[runningheads]{llncs}
\usepackage{graphicx}
\usepackage{microtype}
\usepackage[T1]{fontenc}
\usepackage[utf8]{inputenc}
\usepackage{amssymb}
\usepackage[english]{babel}
\usepackage{float}
\usepackage{enumitem}
\setlist{nosep} 
\usepackage{tabularx}
\usepackage{booktabs}
\usepackage[sorting=none,maxbibnames=9]{biblatex}

\usepackage{csquotes}
\usepackage[table]{xcolor}
\usepackage{authblk}
\usepackage{slashbox}

\usepackage{commath}
\usepackage{url}
\usepackage[colorlinks=false]{hyperref}
\bibliography{mybib.bib}
\usepackage[normalem]{ulem}
\usepackage{xspace,geometry,fancyhdr}
\usepackage{setspace}
\usepackage{comment}
\usepackage[nameinlink,noabbrev]{cleveref}
\usepackage{mathtools}
\usepackage{placeins}
\usepackage{glossaries}
\usepackage{caption}
\usepackage{subcaption}

\makeglossaries
\newacronym{PM}{PM}{Process Mining}
\newacronym{BPM}{BPM}{Business Process Management}
\newacronym{ML}{ML}{Machine Learning}
\newacronym{DM}{DM}{Data Mining}
\newacronym{LHS}{LHS}{Left-Hand-Side}
\newacronym{RHS}{RHS}{Right-Hand-Side}
\newacronym{CNN}{CNN}{Convolutional Neural Network}
\newacronym{DK}{DK}{Deterministically known}
\newacronym{SK}{SK}{Stochastically known}
\newacronym{IM}{IM}{Inductive Miner}

\newcommand{\algName}{\emph{S-ABCC}}
\newcommand{\algNameSpace}{\emph{S-ABCC }}

\begin{document}
\title{Conformance Checking Over Stochastically Known Logs}
%
%
\author{Eli Bogdanov\inst{1} \and
Izack Cohen\inst{2}\orcidID{0000-0002-6775-3256} \and
Avigdor Gal\inst{1}\orcidID{0000-0002-7028-661X}}
\authorrunning{Bogdanov et al.}
%
\institute{Technion -- Israel Institute of Technology, Faculty of Industrial Engineering \& Management, Haifa 3200003, Israel  \and
Bar-Ilan University, Faculty of Engineering, Ramat Gan 5290002, Israel\\}

\maketitle              
\begin{abstract}
With the growing number of devices, sensors and digital systems, data logs may become uncertain due to, {\em e.g.},  sensor reading inaccuracies or incorrect interpretation of readings by processing programs. At times, such uncertainties can be captured stochastically, especially when using 
probabilistic data classification models.
In this work we focus on conformance checking, which compares a process model with an event log, when event logs are stochastically known. Building on existing alignment-based conformance checking fundamentals, we mathematically define a stochastic trace model, a stochastic synchronous product, and a cost function that reflects the uncertainty of events in a log. Then, we search for an optimal alignment over the reachability graph of the stochastic synchronous product for finding an optimal alignment between a model and a stochastic process observation. Via structured experiments with two well-known process mining benchmarks, we explore the behavior of the suggested stochastic conformance checking approach and compare it to a standard alignment-based approach as well as to an approach that creates a lower bound on performance. We envision the proposed stochastic conformance checking approach as a viable process mining component for future analysis of stochastic event logs.

\end{abstract}
\section{Introduction} \label{sec:intro}
Process mining relies on
data that are typically
stored in the form of event logs and collections of traces where each trace is a sequence of events and activities that were created following a process realization. 
Process mining tasks, such as conformance checking, use event logs to achieve their goal ({\em e.g.}, assessing to what degree a process model and an event log conform) of 
improving the process model that generates these logs.

The fourth industrial revolution \cite{schwab2017fourth}, which is bridging our digital and physical worlds, is producing an abundance of event data from multiple sources such as social media networks \cite{sener2018unsupervised}, sensors located within smart cities ({\em e.g.}, the \href{https://biu-vf-project.wixsite.com/biuvfproject}{`Green Wall'} project in Tel Aviv and Nanjing), medical devices and much more. Differently from data within traditional information systems, these data may involve uncertainty due to technical reasons such as sensor inaccuracy, the use of probabilistic data classification models, data quality reduction during processing, and low quality of data capturing devices. Human generated data may be uncertain as well, due to fake news and mediator interventions. 

In this work, we focus on process mining with \gls{SK} event data~\cite{cohen2021uncertain} where the probability distribution functions of the event data are known.\footnote[1]{It is also denoted as `weakly uncertain' event data in the process mining literature; see \cite{pegoraro2020conformance}.} 
By way of motivation, consider a use-case of food preparation processes, captured in video clips that are analyzed by a pre-trained \gls{CNN} to predict activity classes and their sequence within an observed video. To extract the trace of the realized process, one can use the softmax layer of the \gls{CNN} to yield a discrete probability distribution of the predicted activity classes in the observed video. This probabilistic knowledge, in turn, can serve as a basis for an \gls{SK} log. 

Specifically, we develop a conformance checking algorithm over \gls{SK} data. Building on existing alignment-based conformance checking fundamentals, we mathematically define a stochastic trace model, a stochastic synchronous product, and a cost function that reflects the uncertainty of events in a log. Then, we search for an optimal alignment over the reachability graph of the stochastic synchronous product to find an optimal alignment.
The main contributions of this work are:
\begin{enumerate}
    \item We characterize and mathematically define the building blocks for stochastic conformance checking, including a stochastic trace model and a stochastic synchronous product.
    \item We develop a novel conformance checking algorithm between a model and an \gls{SK} trace.
    \item Using publicly available data sets, we evaluate the performance of stochastic conformance checking 
    and highlight unique features of our proposed algorithm.
\end{enumerate}

The rest of the paper is organized as follows. In Section~\ref{sec:Taxonomy}, we develop the model followed by presentation of our stochastic alignment algorithm (Section~\ref{sec:algorithm}). Empirical evaluation of the two is given in detail in Section~\ref{sec:experiments}. The related literature is presented in Section~\ref{sec:literature} and the final section (Section~\ref{sec:conclusion}) concludes the paper and offers directions for future research.

\section{Stochastic Trace Model} \label{sec:Taxonomy} 
\vspace{-5pt}
Uncertain data have recently become a subject of interest among the process mining community~\cite{pegoraro2020conformance,pegoraro2019mining,pegoraro2019discovering}. 
Table~\ref{tab:ProcLog}~\cite{cohen2021uncertain} presents a model/observation classification scheme that is based on the number of models present in a log and whether the log is deterministically or stochastically known.
In this work we focus on 
Case 5, handling a \gls{DK} process model and an \gls{SK} trace, where the decision-maker wishes to identify a conformance measure between the process and the \gls{SK} trace. While the suggested approach can be extended to solve Case 7, we leave this extension as well as other cases for future work.
\begin{table}[H]
	\centering
	\bgroup
\def\arraystretch{1.3}
\begin{tabular}{l c c | c c}
				\midrule
				\hskip.6in \textbf{Model (Data set)} $\rightarrow$ & \multicolumn{2}{l|}{Single process}  & \multicolumn{2}{l}{Multiple processes}\\
				$\downarrow$ \textbf{Observation (Log)} & \gls{DK} & \gls{SK} & \gls{DK}& \gls{SK}\\ \midrule
				Deterministically Known (\gls{DK})    & 1 & 2 & 3 & 4\\
				   \midrule
				Stochastically Known (\gls{SK}) & \cellcolor{yellow}5 & 6   &7 & 8\\
				       \midrule
		\end{tabular}
		\egroup
		\vspace{5pt}
		\caption{Eight cases according to the characteristics of the process and observed log, from \citeauthor{cohen2021uncertain} \cite{cohen2021uncertain}. The present paper focuses on Case 5 (highlighted).}
		\label{tab:ProcLog}
\end{table}
\vspace{-15pt}

Following~\citeauthor{cohen2021uncertain}~\cite{cohen2021uncertain}, we use \gls{DK} to describe a given and known process or event log, which is the common setting in the process mining literature. An \gls{SK} event log has at least one event attribute that can be characterized via a probability distribution. Table~\ref{tab:table2} illustrates an \gls{SK} trace, which we use as the running example throughout the paper. 
\begin{table}[H]
  \centering
    \begin{tabular}{c|c|c|c}
      \textbf{Case ID} & \textbf{Event ID} & \textbf{Activity} & \textbf{Timestamp} \\ 
      \hline
      1 & $e_1$ & $\{A:1.0\}$ & 13-08-2020T12:00  \\ 
      1 & $e_2$ & $\{B:0.2, C:0.8\}$ & 13-08-2020T14:55  \\ 
      1 & $e_3$ & $\{D:0.6, E:0.2, F:0.1, G:0.1\}$ & 15-08-2020T17:39  \\
      1 & $e_4$ & $\{F:1.0\}$ & 15-08-2020T19:47  
      \vspace{5pt}
    \end{tabular}
  \caption{\label{tab:table2} \gls{SK} data, which is aligned with Case 5 in Table~\ref{tab:ProcLog} in~\cite{cohen2021uncertain}.}
  \end{table}
We now introduce our primary notation and related definitions. We consider a finite set of activities $\mathcal{A}$ and a Petri net $N$ with initial and final markings $m_i$ and $m_f$, respectively. The Petri net is composed of finite sets of places $P$, transitions $T$ and flow relations $F$, which are directed edges among places and transitions. Each transition is associated with an activity $a \in \mathcal{A} \cup \tau$ by the labeling function $\lambda: T \rightarrow A^\tau$ $(A^\tau \equiv A \cup \tau)$. $\tau$ is 
a silent activity separate from the other activities in $\mathcal{A}$. 

Differently from a \gls{DK} trace that includes a sequence of activities with probability 1, the activities in an \gls{SK} trace are associated with a probability function ({\em e.g.}, the next transition may be `act1' with probability $p$ or `act2' with probability $1-p$). We reflect the stochastic nature of the traces using a weight function $W: T \rightarrow (0,1)$ that assigns a firing probability to each transition. 

Our modeling approach is inspired by a conformance checking algorithm~\cite{carmona2018conformance} (pp. 125-158) to align a \gls{DK} trace and a model's execution sequence such that the cost of dissimilarities is minimized.      
The algorithm by \cite{carmona2018conformance} 
cannot be used directly with \gls{SK} traces. Our proposed model, however, aims to provide this ability.
In what follows, we assume prior knowledge about alignment-based conformance checking and related definitions ({\em e.g.}, system net, process and trace models, and synchronous product). We refer  interested readers to~\cite{carmona2018conformance} for a thorough description of relevant definitions and methods. 

We start by defining a stochastic trace model.
\begin{sloppypar}
\begin{definition}[Stochastic Trace Model]
Let $A \subseteq \mathcal{A}$ be a set of activities, and $\sigma \in A^*$ a sequence over these activities. A {\em stochastic trace model}, $STN =((P,T,F,\lambda,W),m_i,m_f)$ is a system net such that $P = \{p_0,...,p_{|\sigma|}\}$, $T \in \{t_{11},...,t_{|\sigma|n_{\sigma}}\}$, $F \subseteq (P \times T) \cup (T \times P)$ and $W:T \rightarrow (0,1) \;|\; \sum_{j=1}^{n_i} \, W(t_{ij}) = 1, \quad \forall \, 1 \leq i \leq |\sigma|$ where $n_i$ is the number of \textit{parallel} transitions between place $p_{i-1}$ and $p_i$.  $W(t_{i\cdot})$ is a probability function assigning to each parallel transition $j$ a firing probability. Additionally, let $m_i = [p_0]$ and $m_f=[p_{|\sigma|}]$.
\end{definition}
\end{sloppypar}
\begin{figure}[htpb]
	\begin{center}	
			\includegraphics[width=0.9\textwidth]{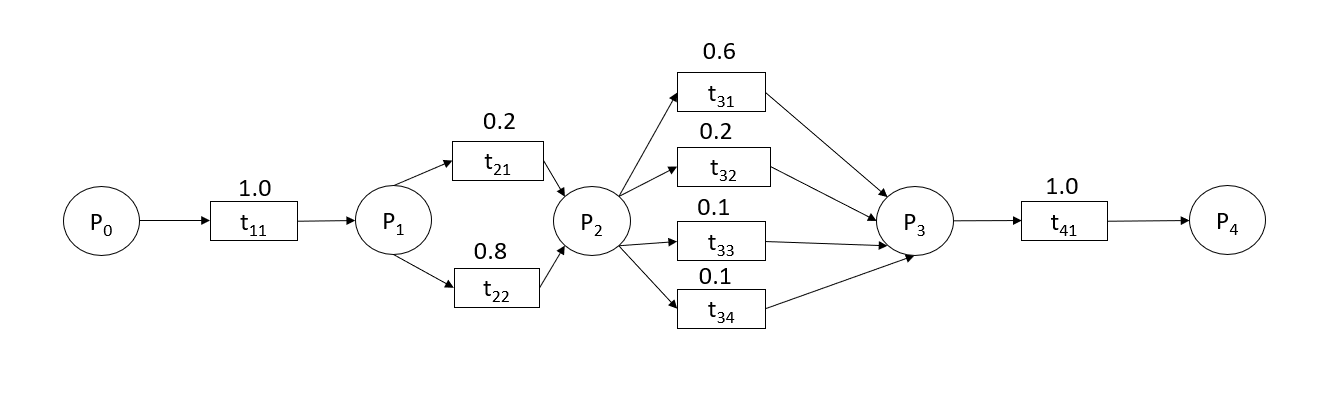}
	\caption{Stochastic trace model illustration}
	\label{fig:stochastic trace model}
	\end{center}
\end{figure}

Figure~\ref{fig:stochastic trace model} offers a visual illustration of a stochastic trace model for our running example from Table \ref{tab:table2}, where transition $t_{11}$ is activity $A$, $t_{21}$ and $t_{22}$ are activities $B$ and $C$, respectively, and so on. The stochastic trace model generalizes a trace model by allowing a place $i$ to have multiple incoming and outgoing edges denoted by $j$, which lead to and from parallel transitions. Each transition has a single outgoing edge from a place and a single incoming edge to a place. Additionally, each transition is associated with a firing probability. For each two places in the Petri net, the sum of firing probabilities of their parallel transitions is 1. 
\vspace{-15pt}
\section{Stochastic Alignment Algorithm}
\label{sec:algorithm}
\vspace{-10pt}
A synchronous product combines process and trace models such that each pair of transitions that are labeled with the same activity are denoted a synchronous transition. Nonsynchronous transitions are represented by pairing an activity with $>>$ and are associated with a cost of 1. 
An optimal alignment between a trace and a model is the execution sequence of the model for which the alignment between the trace and the sequence has the lowest possible cost. De facto, this is an execution sequence of the synchronous product model that produces the lowest cost. 

While deterministic traces have a single execution sequence, for \gls{SK} traces a synchronous product procedure should align multiple model execution sequences with multiple trace execution sequences. We search for the optimal alignment using the reachability graph of the synchronous product. Towards this end, we need to extend the standard version of a synchronous product by including probability functions that capture the \gls{SK} nature of the trace. The probability functions assign a firing probability to each synchronous move of the trace and the model. The probability of the synchronous move is equal to the probability of the same transition in the stochastic trace model as defined next.
\begin{definition}[Stochastic Synchronous Product]\label{def:stoch_synchronous_product} \\
Let $$SN=((P^{SN},T^{SN},F^{SN},\lambda^{SN}), m_i^{SN}, m_f^{SN})$$  be a process model and $$STN=((P^{STN},T^{STN},F^{STN},\lambda^{STN},W^{STN}),m_i^{STN}, m_f^{STN})$$ a stochastic trace model. The stochastic synchronous product $SSN = ((P,T,F,\lambda,W),m_i, m_f)$ is a system net such that:
\begin{itemize}
\item[$\bullet$] $P=P^{SN} \cup P^{STN}$ is the set of places,
\item[$\bullet$] $T=T^{MM} \cup T^{LM} \cup T^{SM} \subseteq (T^{SN} \cup \{>>\}) \times (T^{STN} \cup \{>>\})$ is the set of transitions where $>>$ denotes an $SSN$ transition in which either the model or the trace executes an activity and its counterpart does not, i.e., $>> \notin T^{SN} \cup T^{STN}$, with \\
\hspace*{2mm} $T^{MM} = T^{SN} \times \{>>\}$ (model moves), \\
\hspace*{2mm} $T^{LM} = \{>>\} \times T^{STN}$ (log moves), and \\
\hspace*{2mm} $T^{SM} = \{(t_i,t_j) \in T^{SN} \times T^{STN} \; | \; \lambda^{SN}(t_i) = \lambda^{STN}(t_j)\}$ (synchronous moves). 
\item[$\bullet$] $F=\{(p,(t_i,t_j)) \in P \times T \; | \; (p,t_i) \in F^{SN} \lor (p,t_j) \in F^{STN}\} \cup \{((t_i,t_j),p) \in T \times P \;| \; (t_i,p) \in F^{SN} \lor (t_j,p) \in F^{STN}\}$, \item[$\bullet$] $m_i = m_i^{SN} + m_i^{STN}$,
\item[$\bullet$] $m_f = m_f^{SN} + m_f^{STN}$ and,
\item[$\bullet$] $\forall (t_i,t_j) \in T$ it holds that $\lambda((t_i,t_j)) = (l_i,l_j)$, where $l_i = \lambda^{SN}(t_i)$ if $t_i \in T^{SN}$, and $l_i= >>$ otherwise; and $l_j=\lambda^{STN}(t_j)$, if $t_j \in T^{STN}$, and $l_j = >>$ otherwise. Finally, 
\item[$\bullet$] the probability function $W:T \rightarrow (0,1) \;|\; W^{SSN}(t_i,t_j) = W^{STN}(t_j), \\ \forall(t_i,t_j) \in T^{SSN} : \lambda^{SN}(t_i) = \lambda^{STN}(t_j)$ assigns firing probabilities to transitions of synchronous moves.   
\end{itemize}
\end{definition}

The stochastic synchronous product is a combination of a process model that may yield multiple execution sequences (traces) and a stochastically known trace model that is noisy. Thus, the `real' deterministic trace can be only deduced with probability. The transitions of the stochastic synchronous product are a union of synchronous and nonsynchronous transitions. To combine a process model and a trace in a system net that represents the synchronous product, each pair of transitions that are labeled with the same activity is added as a synchronous transition. Nonsynchronous transitions, which include a process (trace) activity that cannot be matched with the same activity on the trace (model), are paired with $>>$. 

\begin{figure}[htpb]
	\begin{center}	
			\includegraphics[width=0.9\textwidth]{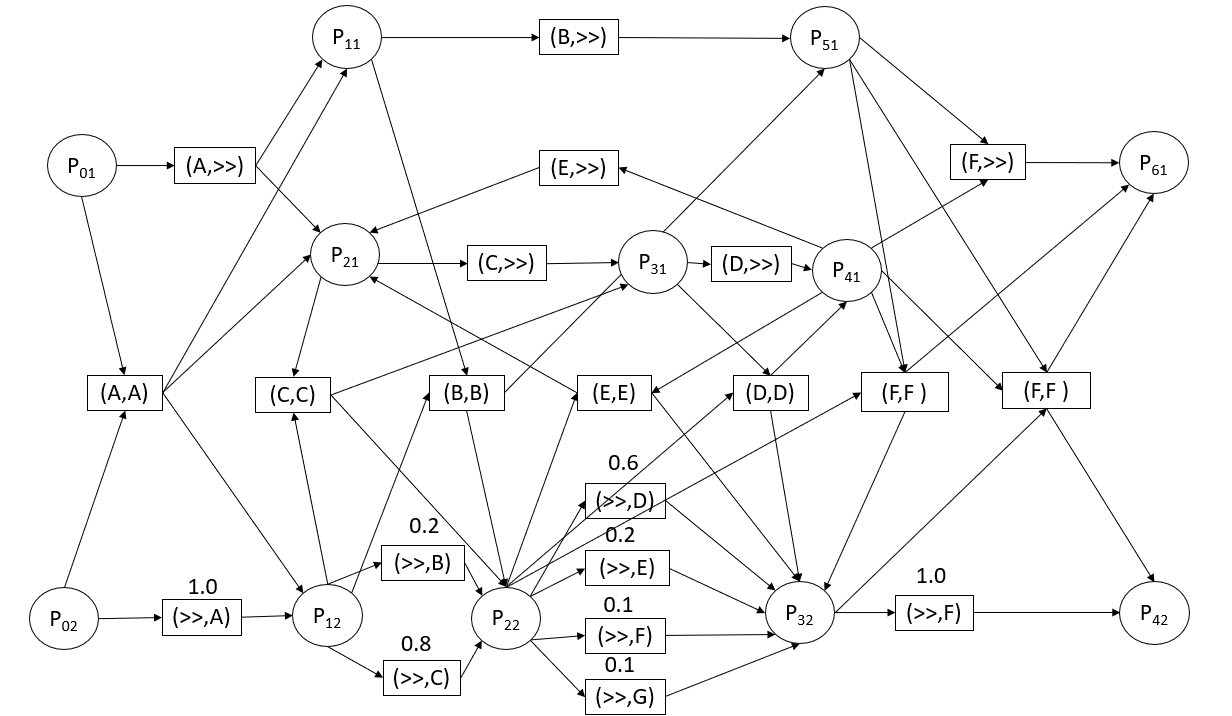}
	\end{center}
	\caption{Stochastic synchronous product illustration}
	\label{fig:stochastic synchronous product}
\end{figure}
Figure~\ref{fig:stochastic synchronous product} illustrates the stochastic synchronous product of a model (its starting place is $P_{01}$) and the stochastic trace of our running example (its starting place is $P_{02}$). The first transition in both the model and the trace is given the label ``activity $A$'' and thus, a new synchronous transition is created---namely, transition $(A,A)$. The original transitions both in the model and the trace are paired with the symbol $>>$ and are added to the new net as well.

We are now ready to introduce our algorithm, \algNameSpace (Stochastic Alignment-Based Conformance Checking), as a solution to the problem of finding the lowest-cost execution sequence of the synchronous product. We observe that this is equivalent to finding the shortest path over the synchronous product's reachability graph, where the sum of costs across path edges is the total path length. 

Given an initial marking $m_i$ of a stochastic synchronous process model $SSN$, we denote the corresponding system net as $N = (P,T,F,\lambda, W)$ and its set of reachable markings as $RS(N)$. The reachability graph of $N$, denoted by $RG(N)$, is a graph in which the set of nodes is the set of markings $RS(N)$ and the edges correspond to firing transitions, where each edge in $RG(N)$ corresponds to a transition of the stochastic synchronous process $SSN$. Formally, an edge $(m_1,t,m_2) \in RS(N) \times T \times RS(N)$ exists, if and only if $m_1[t \rangle m_2$. The shortest path from the initial to the final marking in $RG(N)$ corresponds to the lowest-cost execution sequence of $SSN$. We model the transition probabilities of the \gls{SK} trace in the reachability graph by assigning weights (costs) to the edges as discussed next.

\begin{sloppypar}
Recall that $SSN$ is the stochastic synchronous product of $SN=((P^{SN},T^{SN},F^{SN},\lambda^{SN}), m_i^{SN}, m_f^{SN})$ and a stochastic trace $STN=((P^{TN},T^{TN},F^{TN},\lambda^{TN},W^{TN}),m_i^{TN}, m_f^{TN})$. For every synchronous move, transition $t'=(t_i,t_j)$ in $SSN$ and its corresponding edge $e'$ in $RG(N)$, the cost of $e'$ is calculated by
\begin{equation}
Weight(e') = 1 - e^{1- \frac{1}{W(t')}}, \quad \forall \, t' =(t_i,t_j)\in T^{SSN} \; | \;  \lambda^{SN}(t_i) = \lambda^{STN}(t_j)
\label{eq:cost}
\end{equation}
where $W(t')$ is the firing probability of transition $t'$, and 1 otherwise ($W(e') = 1, \quad \forall \, t' =(t_1,t_2)\in T^{SSN} \; | \; t' \in T^{SN} \times \{>>\} \lor \{>>\} \times T^{STN}$ (model moves or log moves, respectively)). 
\end{sloppypar}

\begin{figure}[htpb]
	\begin{center}	
			\includegraphics[width=0.8\textwidth]{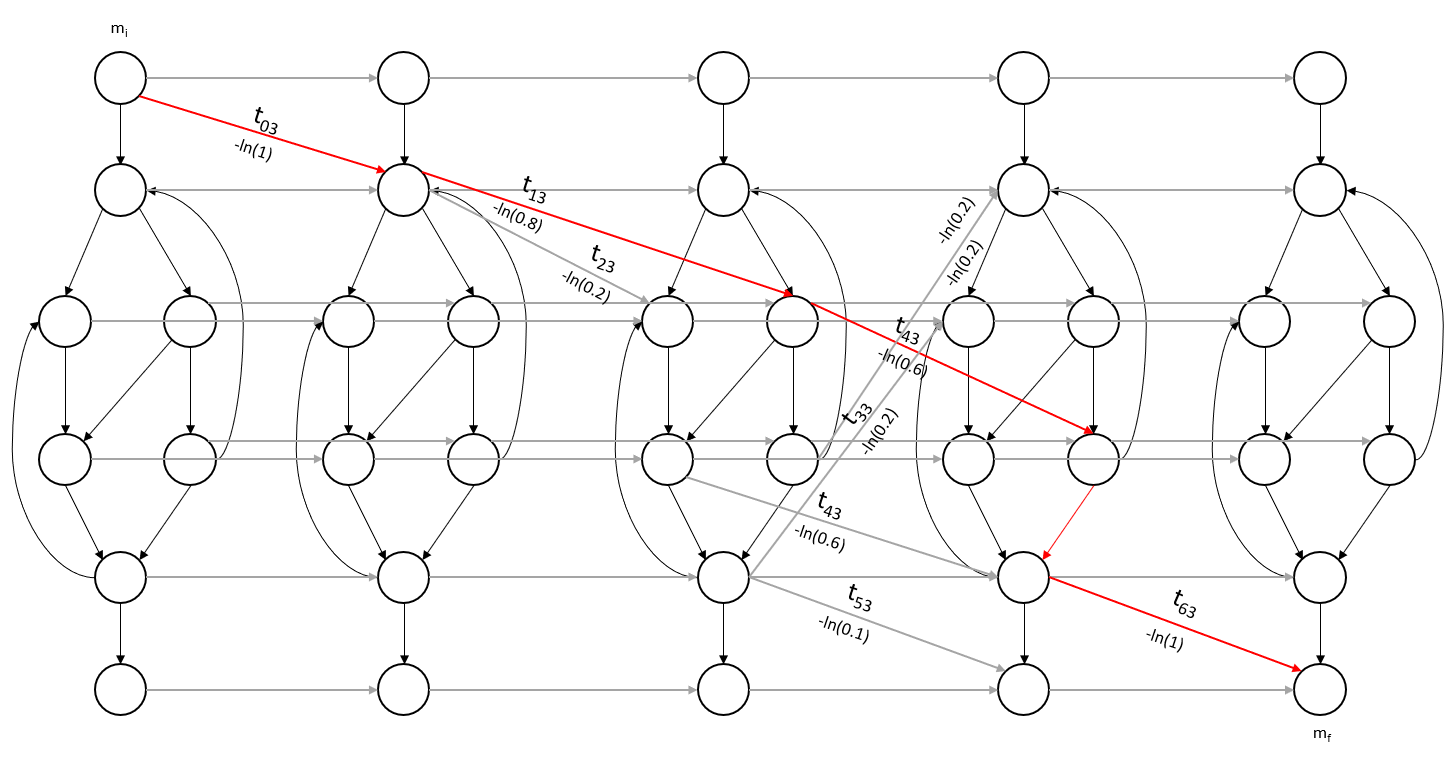}
	\end{center}
	\caption{The reachability graph of the stochastic synchronous product in Figure \ref{fig:stochastic synchronous product}. The red edges mark the optimal path after applying the Dijkstra algorithm.}
	\label{fig:RG}
\end{figure}

The cost function (Eq.~\ref{eq:cost}) transforms firing probabilities into costs. We use a non-linear cost function such that 
each edge $e'$ in the reachability graph $RG(N)$ satisfies the following: $0 \leq Weight(e') \leq 1$. The following property (which proof is omitted due to space considerations) offers guarantees with respect to synchronous moves.
\begin{property}\label{lemma:cost_function}
The cost function (Eq.~\ref{eq:cost}), $f(x)=1 - e^{1 - \frac{1}{x}}$, satisfies the following properties for synchronous moves:
\begin{enumerate}
    \item The cost of an edge in $RG(N)$ approaches $0^+$ as the firing probability of its transition approaches $1$,
    \item it approaches $1$ as the firing probability of the transition approaches $0$, and
    \item $1\leq f(x) \leq 0, \,\, \forall x\in(0,1]$.
\end{enumerate}
\end{property}
For the deterministic setting, the cost of each edge in $RG(N)$ is either 0 or 1 and thus, the deterministic setting can be seen as a special case of our setting with the firing probability of each transition set to 1. 
Given a stochastic synchronous product $SSN$ (Definition~\ref{def:stoch_synchronous_product}) and the cost function (Eq.~\ref{eq:cost}), any shortest path algorithm ({\em e.g.},  Dijkstra~\cite{carmona2018conformance}) can be applied to find the shortest (cheapest) path from the initial to the final markings -- this path corresponds to an optimal alignment between the stochastic trace and the model. To illustrate, Figure~\ref{fig:RG} presents the reachability graph of the stochastic synchronous product in Figure~\ref{fig:stochastic synchronous product} and the shortest path.

\section{Empirical evaluation} \label{sec:experiments}

We evaluate \algNameSpace 
against a standard alignment-based conformance and a lower bound on the conformance cost~\cite{pegoraro2020conformance}. We start with a description of the benchmark data sets (Section~\ref{sec:data}), followed by an explanation of the experiment design (Section~\ref{sec:data_preparation}). We report on the outcome of the empirical evaluation in Section~\ref{sec:results}.
\subsection{The datasets}\label{sec:data} 
We used two publicly available real-world  datasets as a baseline for our experiments: \href{http://icpmconference.org/2019/icpm-2019/contests-challenges/bpi-challenge-2019/}{BPI 2019}  and \href{https://www.win.tue.nl/bpi/doku.php?id=2012:challenge}{BPI 2012}. The BPI 2019  data set contains over 1.5 million events for purchase orders that were collected from a large international coatings and paints company in the Netherlands. The dataset consists of over 250,000 traces relating to 42 activities performed by 627 users. The BPI 2012 dataset consists of about 262,000 events and 13,000 applications for personal loans or overdraft approvals held by a Dutch financial institute. 
\subsection{Data preparation and experiment design}\label{sec:data_preparation}
For each of the data sets, we discovered a baseline model using 15 randomly chosen traces via the \gls{IM} algorithm and the PM4PY package.


Stochastic traces were generated from traces that were not utilized for model discovery. We used 100 traces---15 for the model discovery while the remaining 85 were transformed into stochastic traces. The transformation procedure iterates over each trace, adding parallel transitions with random activities. Both original and added transitions are assigned a firing probability. For example, if the original log contained the following record: $\{CaseID:1,\, EventId:e1,\, Activity:A\}$, a possible corresponding stochastic record after adding transitions with random activities and firing probabilities is $\{CaseID:1,\, EventId:e1,\, Activity:[A:0.4,\, B:0.4,\, C:0.2]\}$. 

We control the following parameters when preparing the stochastic traces.
\begin{itemize}
    \item Number of parallel transitions, $N_t$, varied between 2 and 4. Consider, for example, $N_t=2$, which is two parallel transitions for trace $<A,B,C>$. Then for each of the three events, a second parallel transition is added with an activity that is randomly chosen from the set of activities. 
    \item Value of the firing probability assigned to the original transition 
    in each set of parallel transitions, $P_f$. This parameter is set to one of three values, $P_f \in (0.55,0.75,0.95)$. Since the sum of firing probabilities across each set of parallel transitions equals 1, the leftover probability, $1-P_f$, is randomly split between the other parallel transitions. 
    \item Portion of the uncertain traces, 
    $T_p$. When $T_p=0$, the considered trace is deterministic. We increased the parameter's value in steps of $0.05$. For each iteration in which we increased $T_p$, we selected all the traces from the previous iteration and randomly selected $5\%$ of each trace transitions to be transformed into parallel transitions. The selected $5\%$ only included events without parallel transitions to ensure that when $T_p=1$, $100\%$ of the trace events would have parallel transitions. 
\end{itemize}


We note, in passing, that the stochastic traces that we generated resemble the stochastic output of neural networks for classifying activities in video clips or of sensors for identifying observed signals (for more information, refer to~\cite{cohen2021uncertain}).

\vspace{-10pt}
\subsection{Results}\label{sec:results} 
\vspace{-5pt}

\begin{figure}[ht]
     \centering
     \begin{subfigure}[b]{0.4\textwidth}
         \centering
         \includegraphics[width=\textwidth, height=4.5cm]{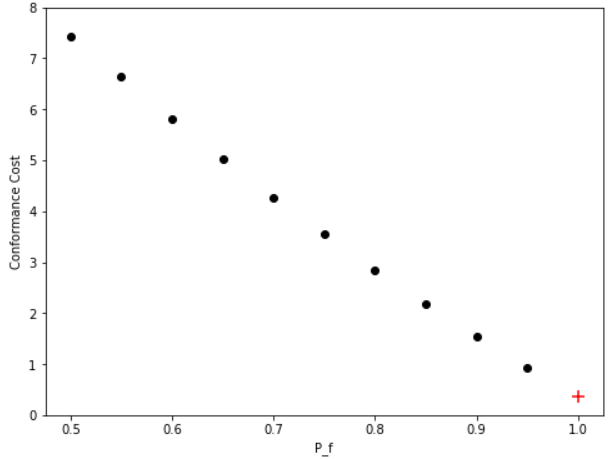}
         \caption{BPI 2012}
         \label{fig:change_activity_labels}
     \end{subfigure}\hfil
     \begin{subfigure}[b]{0.4\textwidth}
         \centering
         \includegraphics[width=\textwidth,height=4.5cm]{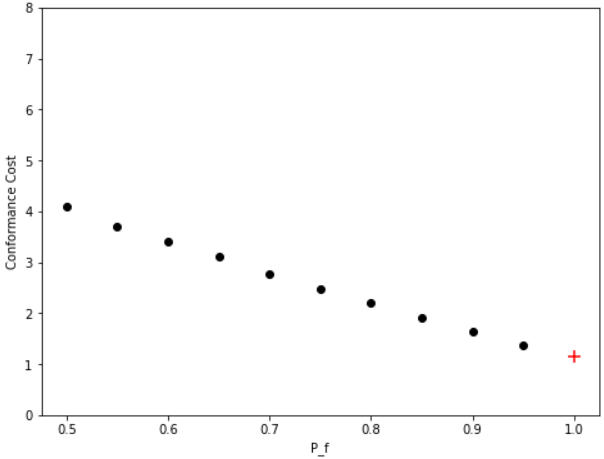}
         \caption{BPI 2019}
         \label{fig:swap_activity_labels}
     \end{subfigure}
        \caption{Average conformance cost as as a function of the firing probability, $P_f$ of the original trace transition. We set $T_p=1$, where each event in the original trace included 2--4 parallel transitions -- $N_t\in(2,3,4)$. The `\textcolor{red}{$+$}' marker corresponds to a deterministic setting.} 
        \label{fig:stochastic_conformance_2012_and_2019}
        
\end{figure}

Figure~\ref{fig:stochastic_conformance_2012_and_2019} demonstrates the sensitivity of the suggested approach to the distribution of the firing probabilities in the sense that changing the firing probability affects the average conformance cost. Specifically, conformance cost decreases with $P_f$ as we get closer to the deterministic setting until it hits the red `+' marker in Figure~\ref{fig:stochastic_conformance_2012_and_2019} in which $P_f=1$. In fact, the suggested model accommodates the deterministic setting in the sense that when assigning $P_f=1$, the suggested model generates the same conformance cost as does conventional alignment-based conformance checking.

Under the suggested model, the optimal alignment carries additional conformance costs compared to its deterministic counterpart due to uncertainty. In a deterministic setting, synchronous moves do not induce a cost, which makes sense since there is only a single trace path. Under an \gls{SK} setting, synchronous moves are associated with a non-negative cost due to uncertainty on the trace path. The extra cost embodies the level of uncertainty for each possible trace realization. Looking at the phenomenon from a different perspective, 
we can say that not accounting for the uncertainty costs would lead to a situation in which as the level of uncertainty increases ({\em e.g.}, by having more transitions in parallel),  the number of possible trace realizations grows and thus we have a greater chance of finding a better conforming trace that is associated with lower conformance costs. This situation is undesirable unless we are seeking a lower bound on the conformance cost (see \cite{pegoraro2020conformance}). 

\begin{figure}[ht]
     \centering
     \begin{subfigure}[b]{0.4\textwidth}
         \centering
         \includegraphics[width=\textwidth, height=4.5cm]{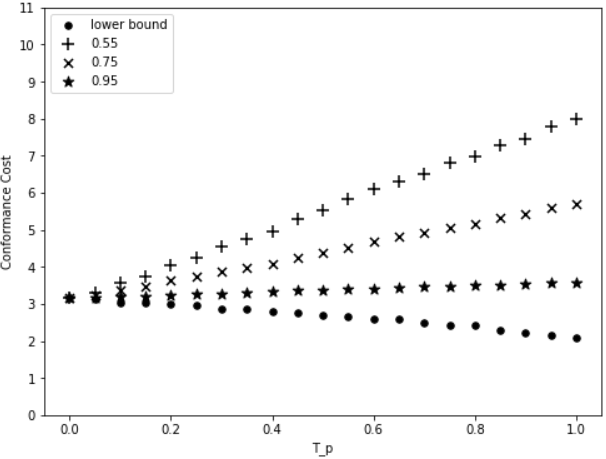}
         \caption{Randomly changing labels for 30\% of the events}
         \label{fig:change_activity_labels_2012}
     \end{subfigure}\hfil
     \begin{subfigure}[b]{0.4\textwidth}
         \centering
         \includegraphics[width=\textwidth,height=4.5cm]{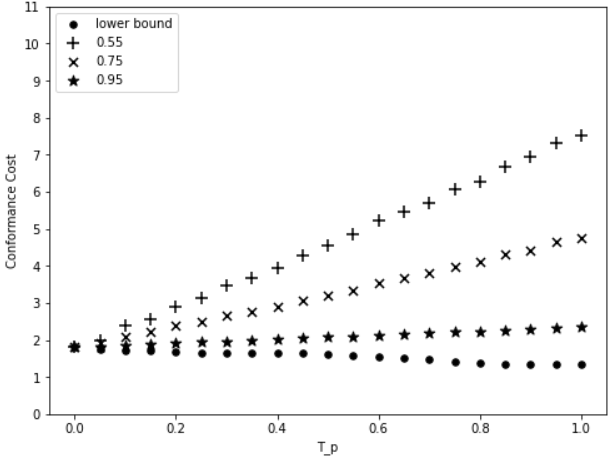}
         \caption{Randomly swapping labels for 30\% of the events}
         \label{fig:swap_activity_labels_2012}
     \end{subfigure} \hfil
     
     \medskip
     \begin{subfigure}[b]{0.4\textwidth}
         \centering
         \includegraphics[width=\textwidth,height=4.5cm]{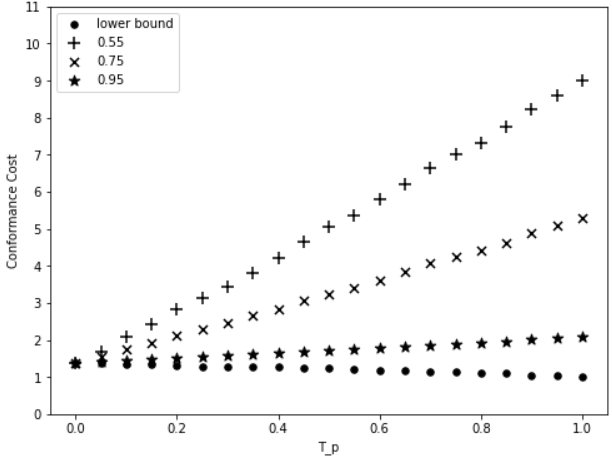}
         \caption{Randomly duplicating 30\% of the events}
         \label{fig:duplicate_activities_2012}
     \end{subfigure}\hfil
     \begin{subfigure}[b]{0.4\textwidth}
         \centering
         \includegraphics[width=\textwidth, height=4.5cm]{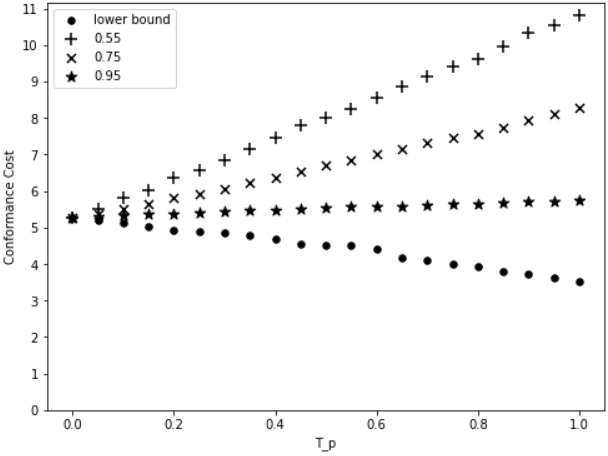}
         \caption{All of the above manipulations}
         \label{fig:all_3_2012}
     \end{subfigure}
        \caption{Average conformance cost as a function of $T_p$, the trace portion with parallel transitions for the four preprocessing modifications as evaluated for the BPI 2012 data set. Different types of markers denote different $P_f$ values and the lower bound; $N_t=2$}
        \label{fig:4_preprocessing_configurations_2012}
\end{figure}

Figure~\ref{fig:4_preprocessing_configurations_2012} 
presents the conformance cost as a function of the stochastic trace portion size for the BPI 2012 data set (results for BPI 2019 showed similar tendencies and are not included due to space considerations).
Inspired by \citeauthor{pegoraro2020conformance} \cite{pegoraro2020conformance}, the original traces were modified prior to adding parallel transitions in one of four ways by: 1) randomly altering the activity label for $30\%$ of the events; 2) randomly swapping $30\%$ of the events with either their successor or predecessor where first and last events in a trace were only swapped with their successor and predecessor, respectively; 3) randomly duplicating $30\%$ of the trace events; and 4) all of the above modifications. After applying a modification, we turn back to the general preprocessing procedure of iteratively adding parallel transitions as detailed in Section \ref{sec:data_preparation}. 
It can be seen in Figure~\ref{fig:4_preprocessing_configurations_2012} that the conformance cost of the \gls{SK} traces increases with $T_p$. On the other hand, the conformance cost of the lower bound, which does not account for probabilities, decreases with $T_p$. This occurs because higher $T_p$ values imply more possible traces and thus additional alignment opportunities while the lower bound does not consider the realization probability of these traces. The result is that the gap, in conformance costs, between the lower bound and the suggested approach that acknowledge uncertainty increases with $T_p$.   

\begin{figure}[ht]
     \centering
     \begin{subfigure}[b]{0.4\textwidth}
         \centering
         \includegraphics[width=\textwidth, height=4.5cm]{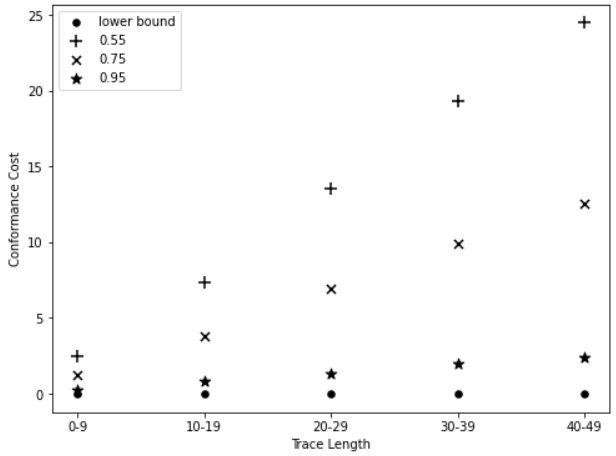}
         \caption{BPI 2012}
         \label{fig:trace_length_2012}
     \end{subfigure}\hfil
     \begin{subfigure}[b]{0.4\textwidth}
         \centering
         \includegraphics[width=\textwidth,height=4.5cm]{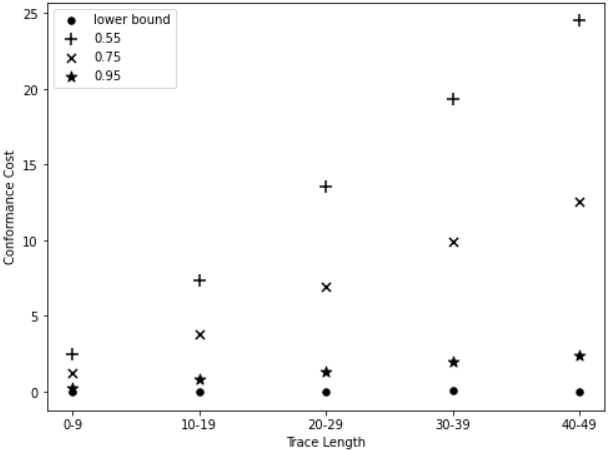}
         \caption{BPI 2019}
         \label{fig:trace_length_2019}
     \end{subfigure}
        \caption{Average conformance cost as a function of the trace length; $N_t=2$, $T_p=1$, $P_f\in(0.55,0.75,0.95)$}
        \label{fig:experiment_4_bpi_2012_2019}
\end{figure}

Next, we evaluated the conformance cost of traces with different lengths. For this, the traces were sorted into groups according to their length, so that group 1 contains traces with a length of 0--9, group 2 contains traces with a length of 10--29 and so on. Following this, we randomly chose three traces from each group (a total of 15 traces) and discovered a model from these traces. Each data point in Figure \ref{fig:experiment_4_bpi_2012_2019} represents the average conformance cost of all the traces that were used for the evaluation, i.e., all the traces within a group excluding the traces that were used for the model discovery.

Figure~\ref{fig:experiment_4_bpi_2012_2019} demonstrates that the conformance cost is increasing with the trace length (apart from the lower bound, for the same reasons explained earlier). The observed behavior follows from the fact that longer stochastic traces have a higher number of possible realizations, which may possibly lead to a better alignment, compared to shorter ones since the number of realizations of a stochastic trace with $T_p=1$ and $N_t=2$ is $2^n$ where $n$ is the length of the trace. We note that the additional cost from synchronous moves outweighs, on average, the reduced cost that may result from a better alignment. 
As can be seen from Figure \ref{fig:experiment_4_bpi_2012_2019} there is no clear trend for the lower bound. This result can be explained similarly to the previous ones: on the one hand, longer traces have longer alignments, which may potentially result in more nonsynchronous moves and a higher conformance cost, but on the other hand, longer traces have a larger realization space when parallel transitions are added (compared to shorter ones as explained before), which may result in a better alignment and thus in a lower overall conformance cost. Such a situation is heavily dependent on the discovered model and its fitness. We leave the study of the degree to which such factors affect the conformance cost in stochastic settings for future work.

 \vspace{-10pt}
\section{Related work} \label{sec:literature}
\vspace{-5pt}
Modeling uncertainty has been introduced in process mining only recently.
Previous studies focused on uncertain data in the sense that some of the data are missing or incorrect and uncertainty is not quantified via any probability distribution. The common approach for dealing with such uncertainty is by preprocessing the event log either by filtering out the affected traces or by repairing existing values~\cite{suriadi2017event, wang2015cleaning, conforti2016filtering, sani2017improving, van2018filtering, conforti2018timestamp}. 
 
To the best of our knowledge, 
uncertainty in event logs was introduced explicitly for the first time in~\citeyear{pegoraro2020conformance} by~\citeauthor{pegoraro2020conformance} \cite{pegoraro2020conformance} who introduced a new taxonomy of uncertainty on the attribute level. At this level, the values of the event attributes are not missing or incorrect but rather appear as a set of possible values and in some cases, the likelihood of each possible value is known or could be estimated. The authors defined two types of uncertainty---namely \textit{strong uncertainty} and \textit {weak uncertainty}. 
The former relates to unknown probabilities between the possible values for the attribute while 
the latter assumes complete probabilistic knowledge in the form of a probability distribution. 
The strong uncertainty setting has been addressed in multiple works. 
A conformance checking technique was proposed by ~\cite{pegoraro2019mining} to compute a lower bound on the conformance cost. \citeauthor{pegoraro2019discovering} \cite{pegoraro2019discovering} described a discovery technique based on uncertain logs that represent an underlying process. In \cite{pegoraro2020efficient} and \cite{pegoraro2020efficientspaceandtime}, the authors proposed an efficient way to construct behavior graphs, which are a graphical representation of precedence relationships among events, for logs with strong uncertain data. By using these graphs, one can discover models from logs through methods based on directly-follows relationships such as the inductive miner \cite{pegoraro2019discovering}. In another recent work by \citeauthor{bergami2021tool} \cite{bergami2021tool}, the authors suggested a technique to compute conformance cost in the setting where the discovered model is assigned probabilities while the traces in the log are deterministic. This work is the first to tackle the problem of conformance checking with \gls{SK} logs.
\vspace{-10pt} 
\section{Conclusion and future work} \label{sec:conclusion}
\vspace{-5pt} 
We developed a conformance checking model for a stochastically known trace in which the probability distribution functions are given. Such a setting may characterize situations in which data logs originate from sensors or probabilistic models. Differently from other conformance checking models, ours explicitly considers the probability values and at the same time accommodates standard (deterministic) alignment-based conformance checking.

When constructing the \algName, in favor of model development, we defined a stochastic trace model and a stochastic synchronous product. Using the stochastic synchronous product and its set of reachable markings, we constructed the corresponding reachability graph. By formulating a bounded non-linear cost function that takes the firing probability as an input, we assigned costs to the edges of the reachability graph that correspond to the stochastic synchronous product. In a final step, we searched over the graph for the shortest (cheapest) path, which represents an optimal alignment where the cost is the conformance cost.
Via structured experiments with two well-known benchmarks, we analyzed the characteristics of 
\algNameSpace and compared it to the deterministic alignment-based conformance checking approach and to a lower bound on the conformance cost. 
On average, the conformance cost of the stochastically known traces converges to their deterministic counterparts as the firing probabilities approach 1. As expected, lower values of firing probability that imply higher uncertainty correspond to higher conformance costs for the same traces. This phenomenon is confirmed when the uncertainty increases due to larger uncertain trace portions. Finally, we observed that conformance costs tend to be higher for longer stochastic traces compared to shorter ones. This occurs because, in general, longer traces may include more synchronous moves that have non-negative costs in the stochastic settings.

This work opens up several interesting future research directions. The first is to use the suggested conformance checking approach to restore the most likely realization from \gls{SK} traces. Possible applications may include improving the accuracy of machine learning classifiers and cleaning errors in datasets.  Another direction is to find both upper and lower bounds on conformance cost. 
Finally, it is worth exploring how different cost functions and search algorithms may affect the performance of \algName. 
\vspace{-15pt}
\printbibliography
\end{document}